\documentclass[conference]{IEEEtran}
\IEEEoverridecommandlockouts
\usepackage{amsmath,amssymb,amsfonts}
\usepackage{algorithmic}

\usepackage{textcomp}
\usepackage{xcolor}

\usepackage[T1]{fontenc}
\usepackage[utf8]{inputenc}
\usepackage[english]{babel}
\usepackage[bookmarks=false]{hyperref}       
\usepackage{url}            
\usepackage{booktabs}       
\usepackage{amsfonts}       
\usepackage{nicefrac}       
\usepackage{microtype}      
\usepackage{multicol}
\usepackage[style=numeric, sorting=none, backend=bibtex]{biblatex}
\usepackage{lipsum}
\usepackage{csquotes}
\usepackage{graphicx}
\usepackage{caption}
\usepackage{subcaption}
\usepackage{multirow}
\usepackage{amssymb}

\usepackage{pifont}

\begin{document}

\title{Recurrence-free unconstrained handwritten text recognition using gated fully convolutional network}

\author{\IEEEauthorblockN{Denis Coquenet, Clément Chatelain, Thierry Paquet}
\IEEEauthorblockA{Normandie University - University of Rouen \\
LITIS Laboratory - EA 4108
\\Rouen, France\\
\{denis.coquenet, clement.chatelain, thierry.paquet\}@litislab.eu}
}

\hypersetup{
pdfauthor={Denis Coquenet, Clément Chatelain, Thierry Paquet}
}

\maketitle
\thispagestyle{plain}
\pagestyle{plain}

\begin{abstract}
Unconstrained handwritten text recognition is a major step in most document analysis tasks. This is generally processed by deep recurrent neural networks and more specifically with the use of Long Short-Term Memory cells. The main drawbacks of these components are the large number of parameters involved and their sequential execution during training and prediction. One alternative solution to using LSTM cells is to compensate the long time memory loss with an heavy use of convolutional layers whose operations can be executed in parallel and which imply fewer parameters. In this paper we present a Gated Fully Convolutional Network architecture that is a recurrence-free alternative to the well-known CNN+LSTM architectures. Our model is trained with the CTC loss and shows competitive results on both the RIMES and IAM datasets. We release all code to enable reproduction of our experiments: \url{https://github.com/FactoDeepLearning/LinePytorchOCR}.
\end{abstract}


\section{Introduction}
Unconstrained offline handwriting recognition consists in analyzing a text line image and outputs a sequence of characters corresponding to that text. This task can be more or less difficult due to variety in size, language or writing style from one writer to another. Moreover, some artefacts can make the work more complex such as overlapping texts, non-character elements or a low resolution of the image. 

This task was first solved using Hidden Markov Model (HMM) and handcrafted features \cite{Ploetz2009}. Nevertheless, those models were limited when dealing with long term dependencies in sequences and they lack discriminative power. To alleviate this problem, some systems combined HMM with other kinds of architecture such as Convolutional Neural Network (CNN) \cite{Bluche2013} or Recurrent Neural Network (RNN) \cite{Frinken2009}. Those hybrid systems (HMM+RNN and CNN+RNN) showed better results over standard HMM.

Then, Long Short-Term Memory (LSTM) Neural Networks appeared to be a key component in this field thanks to their ability to model long range dependencies. However, their large number of parameters and their relatively long training times make them imperfect. More recently, a new trend appeared with the use of recurrence-free models for various tasks including handwriting recognition. They often use gating mechanisms to replace or imitate the LTSM abilities and reduce their amount of parameters by using only convolutional components. They are known as (Gated) Fully Convolutional Networks (G)FCN. 

Whatever the kind of architecture chosen, a language model is generally used as a post-processing stage to increase the performance. However, in this work, we focus only on the optical recognition capabilities, putting any language model or lexicon constraint aside.

The rest of this paper is organized as follows.
Section \ref{related_work} is dedicated to the related works in the field. 
Our GFCN model is presented in details in Section \ref{architecture}. The implementation details with the layers parameters are provided in Section \ref{implementation}.
Section \ref{experiments} is devoted to the experimental study, including datasets description, training details and results compared to state-of-the-art works.

\section{Related Works}
\label{related_work}
There are two major drawbacks with RNN and especially LSTM. LSTM cells require a large number of parameters and the recurrence implies sequential computations, leading to long training and prediction times. In this paper we suggest using a lighter architecture: a Gated Fully Convolutional Network. Thus, we first present state-of-the-art concerning recurrent models and then we go through fully convolutional models.

\subsection{Recurrent Neural Networks (RNN)}
RNN are efficient in a large variety of deep learning tasks including handwritten text recognition. This is especially true for the LSTM that can handle long dependencies in sequences.

The question of using 1D-LSTM or 2D-LSTM on text images is still studied. In \cite{Voigtlaender2016}, a Multi-Dimensional LSTM (MDLSTM) is used to manage dependencies over both horizontal and vertical axis. This leads to increasing learning time and potentially to a huge number of parameters. In \cite{Puigcerver2017}, a 1D-LSTM reaches better results than a 2D-LSTM with less training time but more parameters. More recently, a  2D-LSTM presented in \cite{Moysset2019} showed competitive prediction time and performance over many datasets such as RIMES \cite{RIMES}, IAM \cite{IAM} and more complex ones like MAURDOR \cite{Maurdor}.

The most common architectures are combinations of convolutional layers used as a feature extractor and recurrent layers used as a dependency modeler. Such CNN+LSTM model is used in \cite{Wigington2017} and a more sophisticated version using attention is presented in \cite{chowdhury2018}. \cite{Michael2019} compare the different kinds of attention that are location-based, content-based, penalized, monotonic, monotonic chunkwise and hybrid one. Other works introduce the use of gating mechanism in such architectures \cite{Bluche2017} while others have experimented a sequence to sequence approach \cite{Sueiras2018]}.

Alternatives are studied to add parallelism in RNN computations. Quasi-Recurrent Neural Network (QRNN) \cite{Bradbury2016} uses recurrence over space dimension instead of time and convolution over the time axis. The aim is again to reduce training and prediction times, which may be achieved in a more efficient way with recurrence-free models such as FCN.

\subsection{Fully Convolutional Neural Networks}
Only few works deal with FCN, more specifically in the context of handwritten text recognition.
FCN are only made up of convolutional components, that is to say convolutional and pooling layers. To compensate the loss of the dependency modelling provided by LSTM cells, FCN are usually combined with a selecting mechanism known as gate which acts similarly to gates of LSTM cells. GFCN use gating mechanisms to select the relevant features from one layer to another. Those GFCN have proven their efficiency in many fields of language and image processing such as in language modeling \cite{Dauphin2016}, translation \cite{Gehring2017} and segmentation \cite{Soullard2019}.

Regarding handwritten text recognition, Ptucha et al. \cite{Ptucha2018} presented a CNN reaching competitive results on the IAM dataset. However, this network uses fully connected layers which distinguishes it from FCN. Moreover, it is pre-trained on the RIMES and the NIST datasets ; it is thus not comparable to our work. In \cite{Ingle2019}, a FCN with a gating mechanism is used on the IAM dataset. It uses shared-weights layers to reduce the number of trainable parameters. It reaches state-of-the-art results on that dataset only when using external data or high data augmentation transforming online IAM data into offline data. \cite{Yousef2018} presents a model that heavily uses Depthwise Separable Convolutions \cite{DepthSepConv} to reduce the number of parameters and to have a really deep network. However, it reaches state-of-the-art results on IAM with a huge number of large stacked layers, leading to 26 million of parameters and thus considerable training and prediction times. In a certain way, it then looses the advantage of using a GFCN. We have not succeeded to reproduce their impressive results but it remains a very inspiring work for us.

In this paper we present a deep and light GFCN architecture with few parameters, leading to fast training and prediction while reaching competitive results with respect to state-of-the-art.

\section{Proposed Architecture}
\label{architecture}
The aim of our network is to combine the following features: a good dependency modelling, a small amount of parameters and the only use of convolutional components. These features are respectively obtained by a strong gating mechanism, the use of Depthwise Separable Convolutions and a FCN architecture.
Our model is a deep GFCN which uses the Connectionist Temporal Classification (CTC) loss \cite{Graves2006} for training. This architecture is inspired from our preliminary work on GCNN applied to handwritten text recognition \cite{Coquenet2019}. The key difference between GFCN and GCNN is the removal of the dense layer. It then enables to use input images of variable size. Moreover, we achieved better results with less parameters. In other words, it is only made up of convolutional components: convolutional and pooling layers. No recurrent nor fully connected layers are used.

\subsection{Global description of the model}
We opt for a GFCN because convolutional layers are light components: operations can be parallelized on GPU and they do not require a lot of parameters. Therefore, we can stack many of these layers without impacting memory usage or training and prediction times. Our model contains 22 convolutional layers which enable to reach a receptive field of size (v=196, h=240) where v and h stand for the vertical and horizontal dimension respectively. We mostly use Depthwise Separable Convolutions (DSC) to reduce the number of parameters. As defined in \cite{DepthSepConv}, DSC consists in performing a depthwise spatial convolution followed by a pointwise convolution. The advantage is that the combination of those operations need less trainable parameters than standard convolutions giving comparable results. However, we keep standard convolutional layers at the beginning and at the end of our model because these layers are more crucial to extract features and predict probabilities. Through our experiments DSC turned out to be less efficient when introduced on these layers.

We used another component which is the MaxPooling. It enables to increase the size of the receptive fields while reducing the tensors size and thus the memory usage, preserving the majority of the information. We use it until reducing the vertical dimension to unity while only dividing by four the horizontal dimension so as to keep enough frames to align the horizontal representation with the ground truth using the CTC loss. The most probable symbol (character or blank) is generated for each remaining frame thanks to a softmax activation.

The full model is depicted on Figure \ref{archi}. In order to provide a better view of our model, we have gathered some layers into GateBlocks and ConvBlocks.

\subsection{The gating mechanism}
\begin{figure*}[htbp!]
    \centering
    \begin{subfigure}[b]{\textwidth}
        \includegraphics[width=\linewidth]{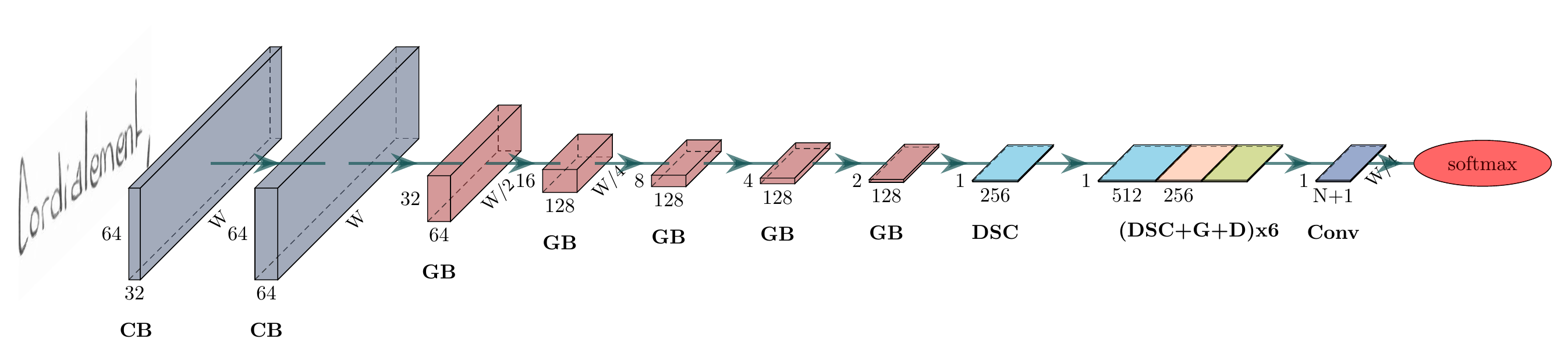}
        \caption{Overview of the GFCN.}
        \label{archi_global}
    \end{subfigure}
    
    \par\bigskip
    \par\bigskip
    
    \begin{subfigure}{\textwidth}
        \centering
        \includegraphics[width=0.7\linewidth]{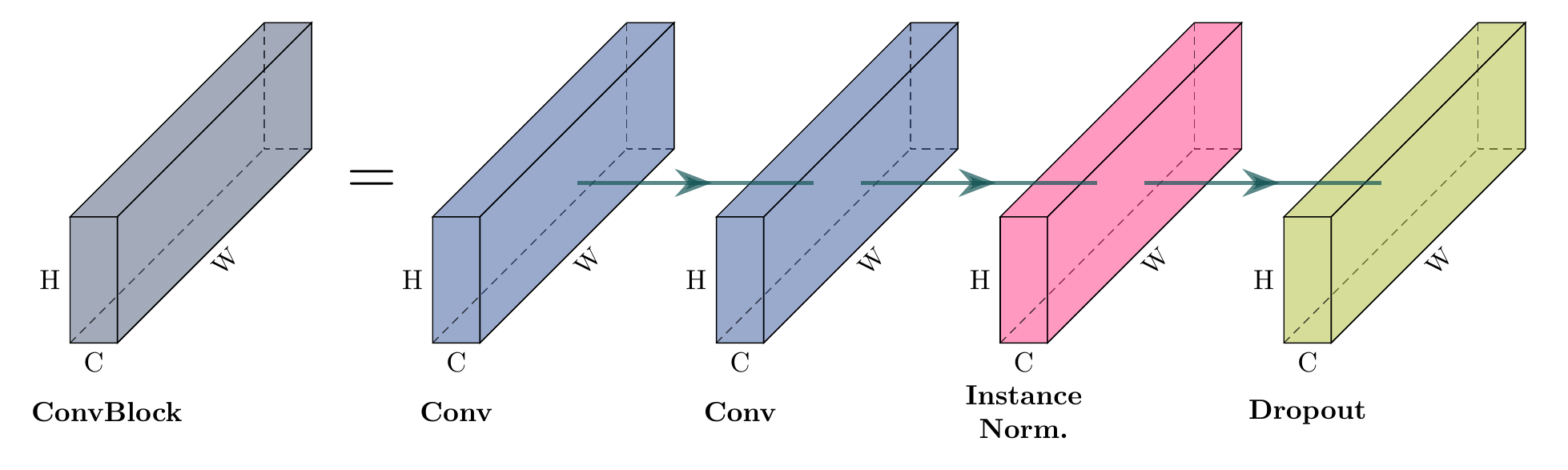}
        \caption{ConvBlock (CB) definition.}
        \label{archi_convblock}
    \end{subfigure}
    
    \par\bigskip
    \par\bigskip
    
    \begin{subfigure}{\textwidth}
        \includegraphics[width=\linewidth]{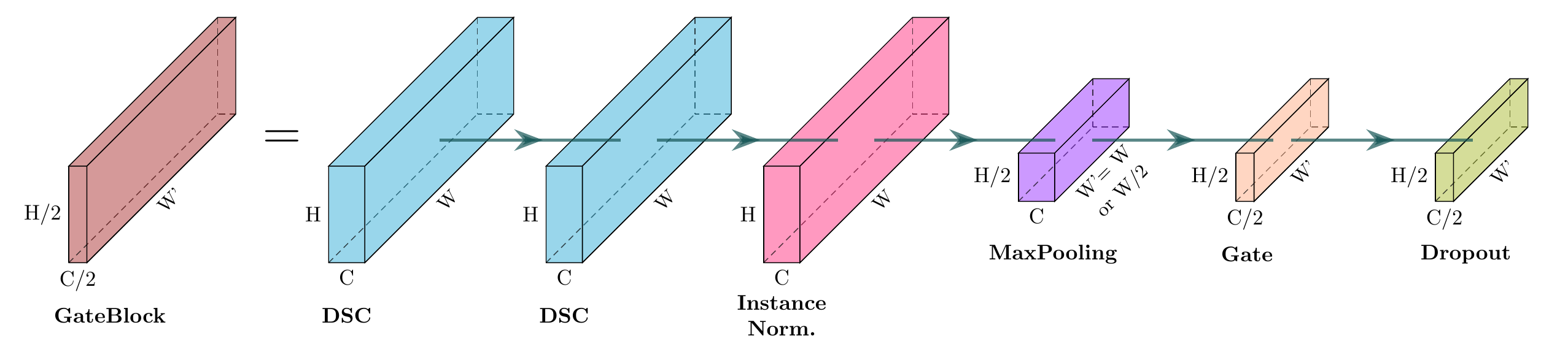}
        \caption{GateBlock (GB) definition.}
        \label{archi_gateblock}
    \end{subfigure}
    
    \par\bigskip
    \par\bigskip
    
    \begin{subfigure}{\textwidth}

        \centering
        \includegraphics[width=0.8\linewidth]{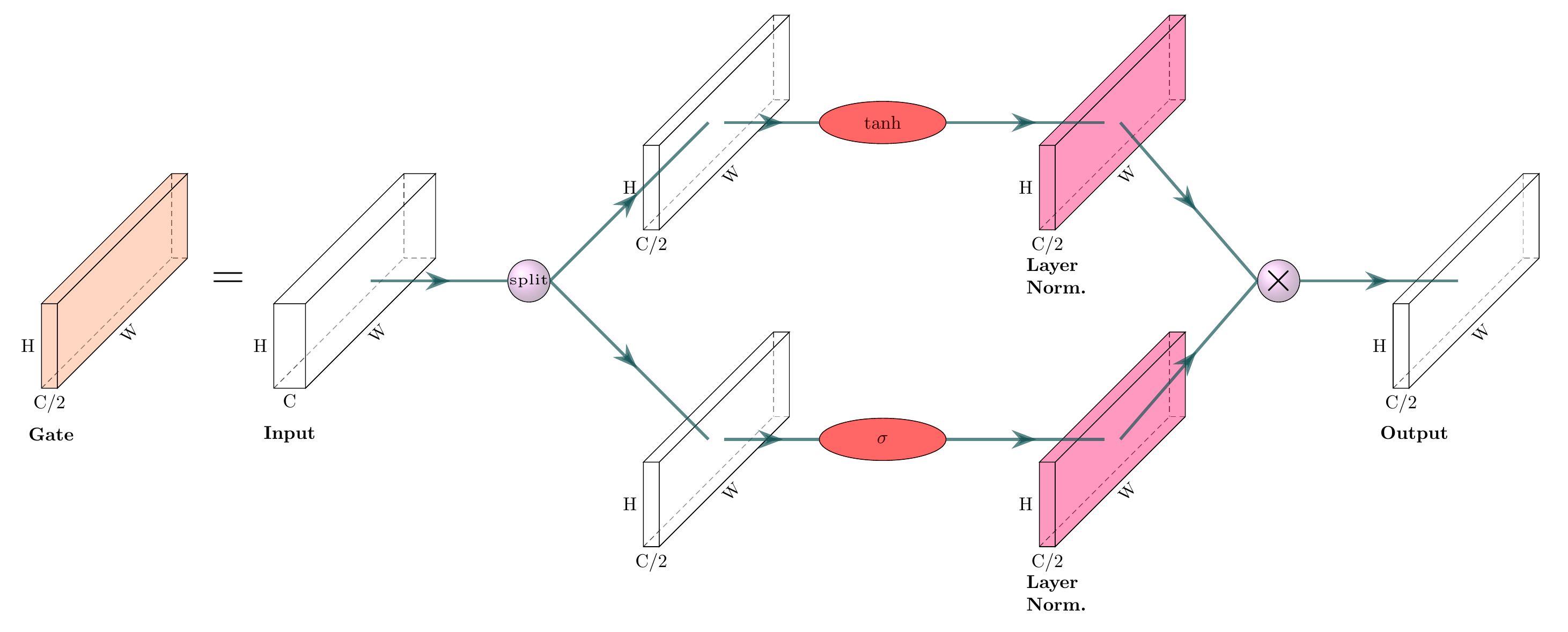}
        \caption{Gating mechanism (G). White elements are just representations of the tensors at a given step (no specific operation is performed).}
        \label{archi_gate}
    \end{subfigure}
    
    \caption{Model definition. Figure \ref{archi_global} presents an overview of the model. The model is made up of some ConvBlocks and GateBlocks which are respectively detailed in Figures \ref{archi_convblock} and \ref{archi_gateblock}. Finally the gate itself is presented in Figure \ref{archi_gate}.}
    \textit{CB: ConvBlock, GB: GateBlock, DSC: Depthwise Separable Convolution, G: Gate, D: Dropout.}
    \label{archi}
\end{figure*}

Our model integrates the concept of Gate which enables to select the relevant features throughout the layers. The gating mechanism is defined as follows: the input tensor is split over the channel axis into two sub-tensors of same dimensions. A $\tanh$ activation is applied to one sub-tensor and a sigmoid to the other. Both are then layer normalized separately before being multiplied element-wise. Thanks to the sigmoid function, the gate can be seen as a selection operator which is used over the layers. A visualization of this mechanism is given on Figure \ref{archi_gate}.

\subsection{Regularization techniques}
Regularization techniques are used to improve the training stability and to reduce overfitting. In this way, the architecture starts with a gaussian noise layer, Instance Normalization is added after the convolutional layers in the ConvBlocks and the GateBlocks, and Dropout is also applied at multiple times. We choose to use the Instance Normalization instead of Batch Normalization which is widely used in the literature because it is mini-batch size independent and it showed great results in other tasks such as stylized image generation \cite{InstanceNorm}. As a matter of fact, since we used small mini-batches, the Batch Normalization is not that stable in our case. This choice is analyzed in Section \ref{norm-choice}.

\section{Implementation details}
\label{implementation}

\subsection{Technical features of the model}
The GFCN takes as input images of 64 pixels in height with variable width. The model starts with 2 ConvBlocks of respectively 32 and 64 filters preserving the original image shape. There are then 5 GateBlocks, the first two with MaxPooling  2$\times$2 kernels, the three following layers with 2$\times$1 kernels. A DSC with 2$\times$1 kernel is then used to reduce the height to 1. A succession of 6 DSC+Gate+Dropout is applied with 1$\times$8 kernels to enlarge the receptive field along horizontal axis. A last convolution with 1$\times$1  kernel  enables us to have a vector of size N+1 for each frame (there are $w/4$ frames). The softmax activation provides probabilities of each character ($N$ being the charset size + 1 for the CTC blank) for each frame which are then used by the CTC loss during training.

\subsection{The ConvBlocks}
A ConvBlock achieves 2 standard convolutions with $C$ filters. Each convolution has the following parameters: $3\times3$ kernel, stride $1\times1$, padding $1\times1$, ReLU activation. Instance Normalization is then applied, followed by Dropout with a probability of $0.4$. A visualisation of a ConvBlock is provided on Figure \ref{archi_convblock}.

\subsection{The GateBlocks}
A GateBlock is defined as a succession of two Depthwise Separable Convolutions (DSC) (C filters, 3$\times$3 kernel, stride 1$\times$1, padding 1$\times$1, each followed by a ReLU activation) followed by Instance Normalization, MaxPooling, Gate and Dropout layers. For the two first GateBlocks, Max Pooling is carried out with a kernel size of 2$\times$2 (then 2$\times$1). This means that we do not divide the width of the image by more than 4. Dropout is applied with a probability of $0.4$. Figure \ref{archi_gateblock} shows a GateBlock in a more visual way.

\section{Experimental study}
\label{experiments}

\subsection{Datasets}
To compare our model to other works, we used two well-known datasets at line level: RIMES and IAM. We used the same split in train, validation and test as the compared works. We do not use any data augmentation technique in the experiments.
Images are resized to obtain a height of 64 pixels conserving their initial width. The input grey level is normalized to zero mean and unit variance.

\subsubsection{RIMES}
The RIMES dataset is made up of 12,723 pages collected in the context of mails writing scenarios. It corresponds to French grayscaled images of handwritings. There are approximately 1,300 writers and the alphabet contains 100 characters. We used the line-segmented version of RIMES that contains 9,947 text lines for training, 1,333 for validation and 778 for test.

\subsubsection{IAM}
The IAM dataset is composed of 1,539 scanned form pages from 657 writers. It corresponds to English handwriting. They are also available as grayscaled images and once more we focus on the line dataset. There are 79 characters in the alphabet and the dataset is split as follows: 6,482 lines for training, 976 for validation and 2915 for test.


\subsection{Training details}
The experiments are carried out with pytorch and the CTC loss was used to train the model.
We used Adam optimizer with an initial learning rate of $10^{-4}$.
Mini-batch of size 2 are used during training and evaluation. Trainings are stopped if no improvement is observed during 50 consecutive epochs.

\subsection{Evaluation}
We use the Character Error Rate (CER) and the Word Error Rate (WER) as metrics to evaluate the performance of the different models over both the validation and the test sets. Both are computed with the Levenshtein distance. We also provide the number of parameters implied in each architecture.

\subsection{Results}

\subsubsection{Normalization choice}
\label{norm-choice}
In a first experiment, we are looking for the best normalization technique for our model. In this respect, we train the model on the RIMES dataset with different normalization layers in place of the instance normalization layers visible in Figure \ref{archi}. We tested the 4 most common normalization techniques namely Batch Normalization \cite{BatchNorm}, Layer Normalization  \cite{LayerNorm}, Instance Normalization  \cite{InstanceNorm} and Group Normalization  \cite{GroupNorm} for a group of size 32. CER are the best CER on the validation set over the first 50, 100, 150 or 200 epochs for a training with a mini-batch size of 2. The training time for one epoch is also given since it can be an important criterion. Times correspond to training carried out on a GPU V100 16 Go. Table \ref{table:norm} sums up this experiment.  

As we can see, the best CER is obtained with Instance Normalization after 100 epochs. Layer Normalization is acting similarly to Instance Normalization but it implies longer training time which makes it less relevant.
Group Normalization performs well too with tiny differences. 
Batch Normalization is the one with the worst CER. This can be explained by the small mini-batches used that lead to a slightly less stable Batch Normalization. As a matter of fact, it is the only normalization that is mini-batch size dependant. The others use the samples in an independent way to compute means and variances. 
Given these results we can conclude that any of these normalization techniques can be considered apart from Batch Normalization which should be preferred with larger mini-batches only.
Instance Normalization turned out to be an option to be considered when dealing with images and small mini-batches; that's why we keep this one in the following experiments.

\begin{table}[!h]
    \centering
    \resizebox{\linewidth}{!}{
    \begin{tabular}{c c c c c c}
    \hline
    \multirow{2}{*}{Normalization} & CER (\%) & CER (\%) & CER (\%) & CER (\%) & Time\\ 
    & 50 epochs & 100 epochs & 150 epochs & 200 epochs & (/epoch)\\
    \hline
    \hline
    Instance & 6.87 & \textbf{5.03} & \textbf{4.47} & \textbf{4.28} & \textbf{8.5 min} \\
    Layer & \textbf{6.75} & 5.04 & \textbf{4.47} & \textbf{4.28} & 15 min\\
    Group (32) & 7.10 & 5.30 & 4.86 & 4.32 & 8.75 min \\
    Batch & 9.6 & 5.7 & 5.4 & 4.8 & \textbf{8.5 min}\\
    \hline
    \end{tabular}
    }
    \caption{Effect of type of normalization for our GFCN with the RIMES dataset (for a mini-batch size of 2). CER is computed on the valid set.}
    \label{table:norm}
\end{table}
\vspace{-0.3cm}

\subsubsection{Results on the RIMES dataset}
Table \ref{table:rimes} shows state-of-the-art results for the RIMES datasets when no language model, lexicon constraints nor data augmentation is used. As we can see our model reaches better results than some recurrent architectures even if it is not at the leading position. Our GFCN reaches a CER of 4.35\% on test set, which is not far from the Puigcerver CNN + 1D-LSTM model that reaches the best CER with 3,3\%. One can notice that, any of the architectures, 1D-LSTM, 2D-LSTM and GFCN have similar CER. 


\begin{table}[!h]
    \centering
    \resizebox{\linewidth}{!}{
    \begin{tabular}{ c c c c c}
    \hline
    \multirow{2}{*}{Architecture} & CER (\%) & WER (\%) & CER (\%) & WER (\%)\\ 
    & validation & validation & test & test\\
    \hline
    \hline
    2D-LSTM \cite{Moysset2019} & 3.32 & 13.24 & 4.94 & 16.03\\
    2D-LSTM-X2 \cite{Moysset2019}& 3.14 & 12.48 & 4.80 & 16.42 \\
    CNN + 1D-LSTM \cite{Moysset2019} & \textbf{2.9} & 11.68 & 4.39 & 14.05\\
    CNN + 1D-LSTM \cite{Puigcerver2017} & 3.0 & & \textbf{3.3}\\
    Ours & 3.82 & 15.60 & 4.35  & 18.01\\
    \hline
    \end{tabular}
    }
    \caption{Comparative results on the RIMES dataset without LM, lexicon, nor data augmentation.}
    \label{table:rimes}
\end{table}
\vspace{-0.3cm}

\subsubsection{Results on the IAM dataset}
The results obtained on the IAM dataset are presented in Table \ref{table:iam}. Once again, no language model, lexicon nor data augmentation are used by the different models.
Our model reaches 7.99\% of CER which is very close to the best recurrent model that achieves 7.73\%.
\begin{table}[!h]
    \centering
    \resizebox{\linewidth}{!}{
    \begin{tabular}{c c c c c}
    \hline
    \multirow{2}{*}{Architecture} & CER (\%) & WER (\%) & CER (\%) & WER (\%)\\ 
    & validation & validation & test & test \\
    \hline
    \hline
    2D-LSTM \cite{Moysset2019} & 5.41 & 20.15 & 8.88 & 29.15 \\
    2D-LSTM-X2 \cite{Moysset2019}& 5.40 & 20.40 & 8.86 & 29.31 \\
    CNN + 1D-LSTM \cite{Puigcerver2017} & 5.1 & & 8.2 & \\
    CNN + 1D-LSTM \cite{Moysset2019} & \textbf{4.62} & 17.31 & \textbf{7.73} & 25.22 \\
    Ours & 5.23 & 21.12 & 7.99 & 28.61 \\
    \hline
    \end{tabular}
    }
    \caption{Comparative results on the IAM dataset without LM, lexicon nor data augmentation.}
    \label{table:iam}
\end{table}
\vspace{-0.3cm}

\subsubsection{Models characteristics comparison}

We now compare the previous models by considering additional features, and not only the performance (CER), that is to say the training time, the prediction time and the number of parameters at stake. We have reproduced the best models seen previously to get those results, for the others (2D-LSTM and 2D-LSTM-X2) we only give the number of parameters. This experiment was conducted on the IAM dataset with images resized to 128 px height and preserving the original width. We add one GateBlock (and thus one MaxPooling layer) to our model in order to be compatible with images of that height. The results are presented in Table \ref{table:comp}. Training time corresponds to the time spent to train a full epoch (train set) and prediction time is the mean time to predict an IAM test set sample, both with a mini-batch size of 2 on a GPU V100 32Go.

\begin{table}[!h]
    \centering
    \resizebox{\linewidth}{!}{
    \begin{tabular}{c c c c}
    \hline
    \multirow{2}{*}{Architecture} & Training time & Prediction time & \multirow{2}{*}{Parameters}\\ 
    & (min/epoch) & (ms/sample)& \\
    \hline
    \hline
    2D-LSTM \cite{Moysset2019} & & & 0.8 M\\
    2D-LSTM-X2 \cite{Moysset2019} & & & 3.3 M\\
    CNN + 1D-LSTM \cite{Puigcerver2017, Moysset2019} & 11.25 & 57 & 9.6 M\\
    Ours & 13.75 & 74 & 1.4 M \\
    \hline
    \end{tabular}
    }
    \caption{Training time, prediction time and number of parameters of different architectures seen in the previous experiments for the IAM dataset with image height of 128px, preserving the original width.}
    \label{table:comp}
\end{table}
\vspace{-0.3cm}

As we can see among the best models, ours is the one with the lowest number of parameters. Training time and prediction time of the CNN + 1D-LSTM \cite{Puigcerver2017, Moysset2019} and of our model are in the same order of magnitude.
This can be explained by the high number of normalization layers used in our model and its depth that counterbalance with the sequential computations of the LSTM layers. 
It has to be noted that 2D-LSTM and 2D-LSTM-X2 architectures performed well too and they do not imply so many parameters. That is especially true for the 2D-LSTM which uses the fewer parameters of all of the reported models. 



\subsubsection{Impact of the receptive field}
This last experiment aims to determine the impact of the receptive field in a GFCN architecture. In this respect, we vary the number of (Depthwise Separable Convolution + Gate + Dropout) used in our model which is fixed to 6 in our baseline. We made this number vary from 1 to 6 ; for each one we kept the best CER on the valid set over the first 100 and 200 epochs and we report the respective number of parameters and receptive field.  The results are presented in Table \ref{table:rf}. 

One can notice that only the horizontal axis is altered by this number since the DSC has a kernel 1$\times$8. Moreover, we double the horizontal receptive field from 1 to 6 (DSC+G+D) which switch from 100 to 240. In the same way, the number of parameters is almost doubled (from 0.7 to 1.4 million).

Considering that characters have an average width of 31 pixels on the IAM test set, the model can see roughly 3 characters when predicting one when using only (DSC+G+D). For our baseline, it can see up to 8 of them. We may assume that this very large receptive field enables to compensate the long-short memory capability of LSTM.

We can easily see a tendency whereby the CER is improved when increasing the number of (DSC+G+D) and thus the receptive field. We may assume that the high CER of example 4 is due to a bad initialization that would be corrected with more epochs. It would be interesting to cross validate these results by repeating this experiment multiple times to consolidate this assumption. Here, doubling the receptive field this way enables to reduce the CER of 2 points at 200 epochs going from 7.83\% down to 5.80\%.

\begin{table}[!h]
    \centering
    \resizebox{\linewidth}{!}{
    \begin{tabular}{c c c c c c}
    \hline
    Number of ending Gates & CER (\%) & CER (\%)  & \multirow{2}{*}{Parameters} & Receptive Field\\ 
    (DSC+G+D) & 100 epochs & 200 epochs & & (h, w)\\
    \hline
    \hline
    6 (baseline) & 6.82 & \textbf{5.80} & 1,375,792 & (196, 240)\\
    5 & \textbf{6.69} & 5.97 & 1,241,904 & (196, 212)\\
    4 & 8.14 & 7.48 & 1,108,016 & (196, 184)\\
    3 & 6.93 & 6.23 & 974,128 & (196, 156)\\
    2 & 7.35 & 6.63 & 840,240 & (196, 128)\\
    1 & 8.30 & 7.83 & 706,352 & (196, 100)\\
    \hline
    \end{tabular}
    }
    \caption{Impact of the receptive field on the IAM dataset. CER is computed over the valid set.}
    \label{table:rf}
\end{table}
\vspace{-0.3cm}

\section{Conclusion}
\label{conclusion}

We have presented one of the first Gated Fully Convolution Networks applied to handwriting line recognition. Our model is a deep but light network which reaches competitive results compared to the best recurrent models without LM, lexicon constraints nor data augmentation. This demonstrates that recurrence-free models should be considered for the task of text recognition. As a matter of fact, regularization techniques (Instance Normalization, Dropout), light operations (Depthwise Separable Convolution) and an efficient selective mechanism provided by the gates enables to have a deep neural network with a large receptive field while preserving a stable and fast training.

\section*{Acknowledgments}
The present work was performed using computing resources of CRIANN (Normandy, France). This work was financially supported by the French Defense Innovation Agency and by the Normandy region.

\vspace{-0.4cm}
\begin{figure}[!h]
\centering
    \includegraphics[height=1.8cm]{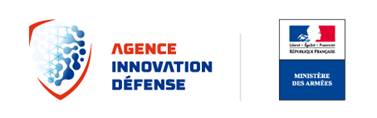}
    ~
    \includegraphics[height=1.8cm]{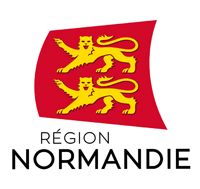}
    \label{data}
\end{figure}
\vspace{-0.5cm}

\printbibliography

\end{document}